\def\year{2022}\relax
\documentclass[letterpaper]{article} 
\usepackage{aaai22}  
\usepackage{times}  
\usepackage{helvet}  
\usepackage{courier}  
\usepackage[hyphens]{url}  
\usepackage{graphicx} 
\usepackage{natbib}  
\usepackage{caption} 
\DeclareCaptionStyle{ruled}{labelfont=normalfont,labelsep=colon,strut=off} 
\usepackage{algorithm}
\usepackage{algorithmic}
\usepackage{amsmath}
\usepackage{url}
\newtheorem{defn}{Definition}
\newtheorem{prop}{Property}
\usepackage{newfloat}
\usepackage{listings}
\floatstyle{ruled}
\newfloat{listing}{tb}{lst}{}
\floatname{listing}{Listing}
\usepackage{fontawesome}
\usepackage{booktabs}
\usepackage{pgfplots}
\pgfplotsset{testbar/.style={
        title=Some title,
        xbar stacked,
        width=10cm,
        axis lines*= left,
        xmin=0,xmax=100,
        ytick = data,
        yticklabels = {Some label},
        tick align = outside,
        bar width=8mm,
        y=10mm,
        enlarge y limits={abs=0.6},
        nodes near coords
    }}
\usepackage{pgf-pie} 

\title{RADAR-X: An Interactive Mixed Initiative Planning Interface Pairing Contrastive Explanations and Revised Plan Suggestions}
\author {
    Karthik Valmeekam$^\dagger$, Sarath Sreedharan$^\dagger$, Sailik Sengupta\footnote{Work done while at Arizona State University.}$^\ddagger$, Subbarao Kambhampati$^\dagger$
}
\affiliations {
    $^\dagger$Arizona State University \\ $^\ddagger$AWS AI Labs\\
    kvalmeek@asu.edu, ssreedh3@asu.edu, sailiks@amazon.com, rao@asu.edu
}

\usepackage{bibentry}
\newcommand{\radarnew}{\texttt{RADAR-X}}
\newcommand{\radar}{\texttt{RADAR}}
\newcommand{\robot}{\mathcal{M}^{R}}
\newcommand{\human}{\mathcal{M}^{H}}
\begin{document}

\maketitle

\begin{abstract}
Decision support systems seek to enable informed decision-making. In the recent years, automated planning techniques have been leveraged to empower such systems to better aid the human-in-the-loop. The central idea for such decision support systems is to augment the capabilities of the human-in-the-loop with automated planning techniques and enhance the quality of decision-making. In addition to providing planning support, effective decision support systems must be able to provide intuitive explanations based on specific user queries for proposed decisions to its end users. Using this as motivation, we present our decision support system \radarnew~that showcases the ability to engage the user in an interactive explanatory dialogue by first enabling them to specify an alternative to a proposed decision (which we refer to as foils), and then providing contrastive explanations to these user-specified foils which helps the user understand why a specific plan was chosen over the alternative (or foil). Furthermore, the system uses this dialogue to elicit the user's latent preferences and provides revised plan suggestions through three different interaction strategies.
\end{abstract}

\section{Introduction}

Proactive decision support systems are a case of human-in-the-loop planning \cite{kambhampati2015human} where the human is responsible for making the decisions and is supported by an automated planning system in complex decision-making scenarios. In scenarios like Navy mission planning, where the commander has to keep track of a lot of information and might lose situational awareness given the complexity of the situation, decision support systems would be useful to provide timely support and help the commander regain situational awareness. In fact, such systems have been shown to aid the user in making faster and better decisions \cite{grover2020radar}. Given that the human (whom we assume to be an expert) is responsible for the final plan in this mixed-initiative setting, a key aspect required for the success of this synergy is to support the user's requirement for explanations, especially when the suggestions made by the system are not acceptable to the user. While previous works on decision support systems \cite{grover2020radar,mishra2019cap} leverage technologies developed in Explainable AI Planning (XAIP) \cite{chakraborti2017plan,sreedharan2018handling,sreedharan2019can}, the participation of the user in explanatory dialogue is limited; \radar~\cite{grover2020radar} provided minimally complete model reconciliation explanations (presented in \cite{chakraborti2017plan}) as and when required but the explanations were not based on specific user queries. This can result in the generated explanations being verbose, making them incomprehensible to the decision-maker. To avoid such situations, the system should let the user drive the dialogue and provide explanations based on the user's query. In this regard, we propose \radarnew, an extension of the \radar~system \cite{grover2020radar}, that supports interactive contrastive explanations \cite{miller2018contrastive} and uses it as the main vehicle for the interaction between the system and the user. We enable users to specify alternatives (referred to as foils) to a plan suggested by the system and ask for explanations that cater to the specified foil. Moreover, we look at the foils as a specification of the user's latent preferences and use that interpretation to come up with refined plan suggestions. We have also made interface improvements that aid in smoothening the interaction process.
\begin{figure}[t]
    \centering
     \includegraphics[keepaspectratio=true, width=\columnwidth]{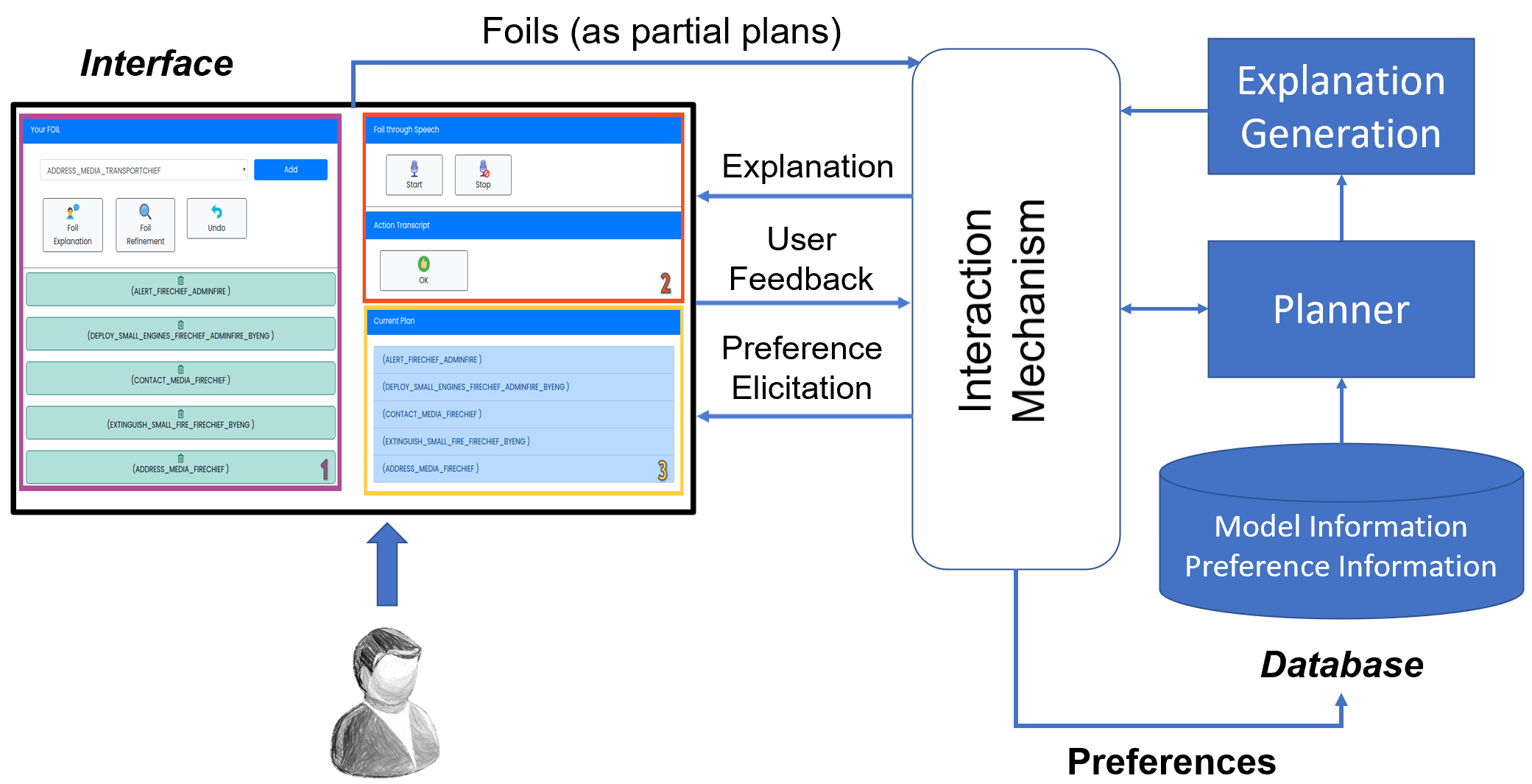}
    \caption{Overview of the \radarnew~system where users can raise foils and ask for contrastive explanations and revised plan suggestions. The system provides these suggestions by eliciting the latent preferences of the user through three different interaction strategies, namely, (1) the nearest plan approach (2) the conflict sets approach and (3) the plausible sets approach.}
    \label{fig:radarx_overview}
\end{figure}

To generate explanations based on user-specified foils, we introduce the idea of a {\em Minimally Contrastive Explanation (MCrE)}, that extends the idea of model reconciliation explanations  \cite{chakraborti2017plan} to answer explicit contrastive queries. We introduce a model-space search and approximation capable of generating MCrE for cases where user foils are represented as partial-plans. A partial-plan can informally be seen as a set of actions with ordering constraints that represent a set of potential solutions to a planning problem \cite{kambhampati1995planning}. Along with providing foil-based explanations, inspired from ideas in iterative planning \cite{smith2012planning}, we consider the case of proactive preference elicitation where plan suggestions are refined based on the specified foil, which are an indication of the user's latent preferences. We look at three different interaction strategies through which plan suggestions are refined-- (1) we develop a novel encoding to generate the nearest plan to the specified foil which implies using the largest possible part of the foil, (2) we employ a search through the space of subsets of the foil to find sets containing conflicting actions and present these sets for the user to resolve, thereby eliciting their preference to generate a plan suggestion and (3) we look to present all the maximal plausible subsets of the foil to the user as options to choose from and use that selection to provide a plan that the user prefers.

In this paper, we start by giving a background of the explanations paradigm that we will leverage. Then, we provide an overview of the implemented interface and a detailed illustration of the domain. With the help of the use-case, we showcase the two technical problems that can be addressed by our system. Finally, we discuss the evaluation conducted for all the aspects of the system and conclude with the path forward for operational deployment.

\section{Background}

In this section, we provide a quick overview of topics in automated planning, necessary to understand the proposed techniques.

A \textbf{Classical Planning Problem} can be described as a tuple $\mathcal{M}=\langle\mathcal{D},I,G\rangle$, consisting of a domain $\mathcal{D}=\langle F,A\rangle$ where $F$ is a finite set of fluent symbols that define a state $s \subseteq F$ and $A$ corresponds to a finite set of actions and $I,G~(\subseteq F)$ represent the initial and goal states. An action $a \in A$ is associated with a cost $c_a$, a set of preconditions $\textit{pre}(a) \subseteq F$ and a set of effects $\textit{eff}(a) \subseteq F$. These effects can be further separated into a set of add effects ${\textit{eff}}^{+}(a)$ and a set of delete effects ${\textit{eff}}^{-}(a)$. The action $a \in A$ can be represented as a tuple $\langle c_a, \textit{pre}(a), {\textit{eff}}^{+}(a), {\textit{eff}}^{-}(a)  \rangle$ and can only be executed in a state $s$ if $s \models \textit{pre}(a)$ i.e., $\delta_{\mathcal{M}}(s,a) \models s \cup \textit{eff}^{+}(a)~ \backslash~ \textit{eff}^{-}(a) ~\textit{if}~ s \models \textit{pre}(a); ~\textit{else}, \delta_{\mathcal{M}}(s,a) \models \perp$ where $\delta_{\mathcal{M}}(.)$ is the transition function. The solution to such a problem is a plan $\pi$ defined as a sequence of actions $\langle a_1, a_2,..., a_n \rangle$ such that
$\delta_{\mathcal{M}}(I,\pi) \models G$ and $\delta_{\mathcal{M}}$ 
here is a cumulative transition function given by $\delta_{\mathcal{M}}(s,\langle a_1, a_2,..., a_n \rangle) = \delta_{\mathcal{M}}(\delta_{\mathcal{M}}(s,a_1), \langle a_2,..., a_n \rangle)$. A sequence of actions that has an unmet precondition and thus, cannot achieve the goal has cost $\infty$. On the other hand, the cost of a plan $\pi$ is the sum of the costs of all the actions present in the plan and is given by $C(\pi,\mathcal{M}) = \sum_{a \in \pi} c_a$. The cost of the optimal plan $\pi^{*}$ is denoted as $C_{\mathcal{M}}^{*}$ where $\pi^{*} = \arg \min_{\pi} \{C(\pi,\mathcal{M}) \forall~\pi$ where $\delta_{\mathcal{M}}(I,\pi)\models G\}$.

In classical planning, the human is considered to have the same planning model and reasoning capabilities as the planner but often, the human's understanding may significantly differ from that of the planner. Thus, we can view this as a \textbf{Multi-Model Planning (MMP)} scenario where $\mathcal{M}^R = \langle \mathcal{D}^R,I^R,G^R  \rangle$ is the planner's model of the planning problem and $\mathcal{M}^H = \langle \mathcal{D}^H,I^H,G^H  \rangle$ is the human's understanding of the same. The difference between these two models, thus, becomes a key factor in the explanation setting. The system tries to achieve common ground with the human by bringing the human's model closer to the system's model through explanations in the form of model updates. This is formalized as a Model Reconciliation Problem in \cite{chakraborti2017plan}.

\textbf{A Model Reconciliation Problem} (MRP), as defined in \cite{chakraborti2017plan}, can be represented using the tuple $\langle \pi^*, \langle \robot,\human \rangle \rangle$ where $\pi^*$ is the optimal plan in $\robot$ ($C(\pi^*,\robot)=C^*_{\robot})$.
MRP is constructed as a model-space search whose solution is considered as an explanation consisting of a set of model updates. Of the four types of explanations defined in \cite{chakraborti2017plan}, \radar~considers the minimally complete explanations \cite{grover2020radar}, which we now define.

A \textbf{Minimally Complete Explanation (MCE)} is the minimal set of relevant information that is provided to the human to explain the optimality of the plan. The objective here is to find the minimum number of differences between the human's model ($\human$) and the planner's model ($\robot$) such that the plan in the planner's model is optimal in the updated human's model.  It is given by 
$\mathcal{E}^{MCE} = \arg \min_\mathcal{E} |\Gamma{(\widehat{\mathcal{M}})}\Delta\Gamma{(\human)}|$
with $C(\pi^*, \widehat{\mathcal{M}}) = C^*_{\widehat{\mathcal{M}}}$ where $\Gamma(\mathcal{M})$ denotes a mapping function that represents a planning problem $\mathcal{M}=\langle\langle F,A\rangle,\mathcal{I,G}\rangle$ as a state in the space of models and $\widehat{\mathcal{M}}$ denotes the model obtained by incorporating the information present in $\mathcal{E}^{MCE}$ into the human model ( $\widehat{\mathcal{M}} = \human + \mathcal{E}^{MCE}$). For a more detailed discussion on MCE we refer the reader to \cite{chakraborti2017plan}.



MCE, although time-consuming to compute, tries to decrease the human's cognitive load by reducing the amount of irrelevant information provided as part of an explanation. Thus, in this work, we try to adapt the objective of MCE for generating contrastive explanations. Before we elaborate on that, we first delve into the interface of \radarnew.

\begin{figure*}[t]
    \centering
     \includegraphics[keepaspectratio=true, width=0.7\textwidth]{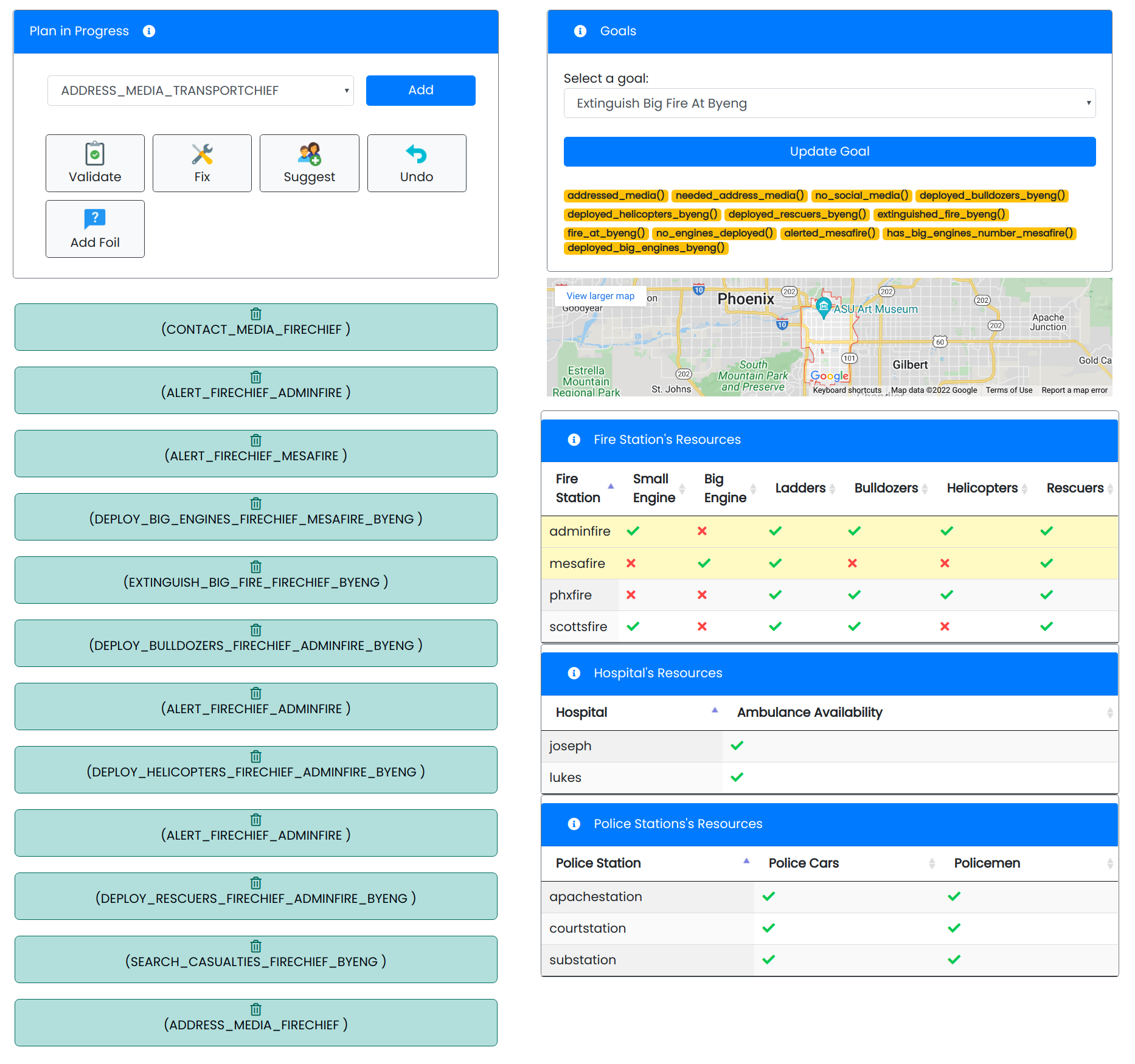}
    \caption{The interface of \texttt{RADAR-X} with various decision supporting functionalities for a human commander making plans in response to a fire. }
    \label{fig:radar}
\end{figure*}
\section{\radarnew}
In \cite{grover2020radar}, the authors consider a fire fighting scenario where \radar~is helping a fire-fighting chief, along with several other authorities, build a plan to extinguish a fire in Tempe, Arizona, USA. This scenario was represented as a classical planning problem in the PDDL \cite{mcdermott1998pddl}. We use the same domain to illustrate the capabilities of \radarnew. We assume that the system has a model of the task ($\mathcal{M}^R = \langle \mathcal{D}^R,I^R,G^R  \rangle$) that may be different from the human's model ($\mathcal{M}^H = \langle \mathcal{D}^H,I^H,G^H  \rangle$) but $\human$ is known to the system $R$ beforehand. Also, we assume $I^H = I^R$ and $G^H = G^R $. 

\begin{figure*}[t]
    \centering
     \includegraphics[keepaspectratio=true, width=\textwidth]{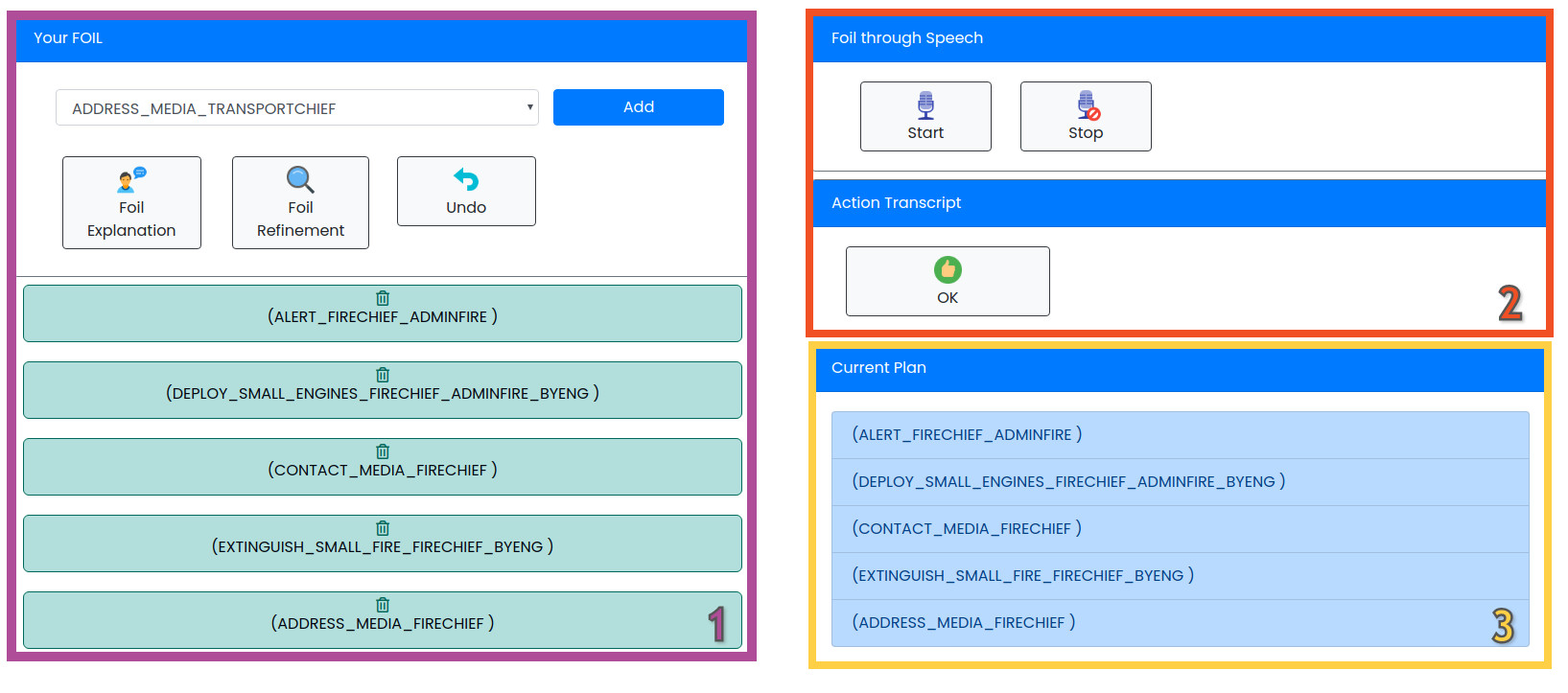}
    \caption{{{\em Add foil} page where users can add actions to their foil and ask for explanations or refined plan suggestions based on the specified foil.}}
    \label{fig:add_foil}
\end{figure*}
\subsection{Overview of the Interface}
\radarnew~seeks to extend the capabilities of \radar~by allowing the user to engage in an explanatory dialogue with the system. Once a plan is suggested by the system, users can utilize the {\tt Add Foil} button present in the {\tt Plan in progress} panel shown in Figure \ref{fig:radar} and are redirected to a new page shown in Figure \ref{fig:add_foil}. On this page, users can provide their foil that is a set of specific actions and ordering constraints over these actions; it may be a partial plan. Let us look at the three panels presented on this new page (as shown in Figure \ref{fig:add_foil}).

\begin{enumerate}
    \item \textbf{Foil Panel}: This is the pivotal panel of \radarnew. This panel allows users to specify foils; it provides the user with the ability to (1) add/delete actions from the foil, (2) change the ordering of the actions present in the foil and (3) choose from options when the foil cannot result in a feasible plan (such as asking for explanations or refined plan suggestions). These options in the panel will be our primary focus in the upcoming sections.
   
    \item \textbf{Foil through Speech}: 
    \begin{itemize}
        \item \textbf{Speech Panel:} Instead of choosing actions from a drop-down list in the Foil panel, this panel facilitates users to specify foils using natural language. For now, we expect the user to say sentences of the form, ``Why contact media firechief and why not contact media transportchief?'' to replace the {\small  CONTACT\_MEDIA\_FIRECHIEF} action in the suggested plan with the expected action of {\small CONTACT\_MEDIA\_TRANSPORTCHIEF}. We have used existing speech-to-text technologies \cite{GoogleSTT} to facilitate such an interaction.
   
    \item \textbf{Action Transcript Panel:}  Once the user specifies the foil using natural language, this panel displays the transcript of their speech; it seeks confirmation before replacing the present plan with the specified foil. This is mostly done to ensure that the inaccuracies in speech recognition technology do not result in the addition of unidentifiable actions. For now, we assume that complete action names are specified when this modality is used and believe that accurate parsing of a user's unrestricted language to foils will be an interesting and helpful future work.
    \end{itemize}

    \item \textbf{Current Plan Panel}: This panel acts as a reference point for the user. While the user works on the specified foils, the panel lists the original plan suggested by the system.
\end{enumerate}
\subsection{Use-Case Description}

As mentioned earlier, the planning problem considered requires a fire-chief, with a decision support system by their side, to come up with a plan for extinguishing a fire in the city of Tempe. In this domain, there are two possible goals-- extinguishing a big fire {\em vs.} extinguishing a small fire. Let us assume that the goal of the user here is to extinguish a small fire. When \radarnew~suggests a plan (as shown in the {\em Current Plan} panel in Figure \ref{fig:add_foil}) that does not meet the expectations of the user (due to differences between $\robot$ and $\human$), the user can specify a foil (that may be a partial plan) and ask for explanations. In our example, there are two specific parts in the suggested plan that are incongruous with the user's expectation. First, the user wants to send out information via a social media post (using {\small SEND\_SOCIAL\_MEDIA}) alongside addressing the media themselves (using {\small ADDRESS\_MEDIA\_FIRECHIEF}), but they are not aware that once posted on social media, the media will pick up the news from these forums, thus not requiring the fire-chief to separately address them. Technically, this is expressed as a delete effect of {\small SEND\_SOCIAL\_MEDIA} called {\small \tt NO\_SOCIAL\_MEDIA} that is also a precondition for {\small ADDRESS\_MEDIA\_FIRECHIEF} in $\robot$. This delete effect is missing in $\human$, leading them to believe that both are possible together.  Second, the user expects that big engines should be deployed along with small engines, i.e. the actions {\small DEPLOY\_SMALL\_ENGINES\_FIRECHIEF\_ADMINFIRE\_BYENG} and {\small DEPLOY\_BIG\_ENGINES\_FIRECHIEF\_MESAFIRE\_BYENG} should both be present in the plan. Here, the user is unaware that both types of engines cannot be deployed together and there is a delete effect of {\small DEPLOY\_SMALL\_ENGINES\_FIRECHIEF\_ADMINFIRE\_BYENG} called {\tt NO\_ENGINES\_DEPLOYED} that is a precondition for {\small DEPLOY\_BIG\_ENGINES\_FIRECHIEF\_MESAFIRE\_BYENG}.
Therefore, when \radarnew~presents the plan suggestion, the user does not expect that only one of each of the expected action pairs, i.e. {\small ADDRESS\_MEDIA\_FIRECHIEF} and {\small DEPLOY\_SMALL\_ENGINES\_FIRECHIEF\_ADMINFIRE\_BYENG}, appears in the suggested plan. The user thus raises a foil $\pi'$ containing the four actions in the given order.

$\pi' = $ 
{\scriptsize
\begin{itemize}
    \item[$-$] DEPLOY\_SMALL\_ENGINES\_FIRECHIEF\_ADMINFIRE \_BYENG
    \item[$-$] DEPLOY\_BIG\_ENGINES\_FIRECHIEF\_MESAFIRE\_BYENG
    \item[$-$] SEND\_SOCIAL\_MEDIA\_BYENG\_BYENG
    \item[$-$] ADDRESS\_MEDIA\_FIRECHIEF
\end{itemize}
}

Note that in this setting, the actions {\small SEND\_SOCIAL\_MEDIA} and {\small DEPLOY\_BIG\_ENGINES\_FIRECHIEF\_MESAFIRE\_BYENG} can also be viewed as the latent preferences of the user which are explicitly elicited when the user raises the foil to ask for explanations. We will now look at how \radarnew~generates (1) explanations to refute the given foil and (2) elicit the preferences of the user through three different interaction strategies.

\subsection{Supporting Contrastive Explanations}
A \textbf{Contrastive Explanation} answers the questions ``Why P and not Q?" where P is the fact (the suggested plan) being explained and Q is the foil (the alternative proposed by the explainee) \cite{miller2019explanation}. 
In \cite{chakraborti2017plan}, the explanation techniques used in \radar~\cite{grover2020radar}, the explanation answers the question ``Why $\pi$?'' where $\pi$ is the suggested plan. Although this can be viewed as an implicit contrastive query, as in ``Why $\pi$ as opposed to any other plan $\pi' (\neq \pi)$?'', it might make the explanation unnecessarily verbose when the user's expected set of plans is much smaller than the set of all optimal plans. Therefore, in \radarnew, we let the user specify their expected set of plans as partial foils and empower the system to generate focused explanations that establish how the current plan compares against their specified set of plans.

We focus on scenarios where the mismatch between the suggested plan and the foil, which is presented as a partial plan, arise due to model mismatch. 
A case for explanation arises when the specified foil (1) cannot be part of a (valid) plan in $\robot$ or (2) is part of a plan but that plan is sub-optimal or costlier than the optimal plan suggested by the system. Thus, the explanation here has to update the human model so they can correctly evaluate the set of plans in question.
\begin{defn}
Given the models $\human$, $\robot$, and a set of plans expected by the human $\hat{\Pi}^H$, a set of model updates $\mathcal{E}^{con}$ is said to be a {\em Minimally Contrastive Explanation (MCrE)}, if
\begin{align}
    \mathcal{E}^{con} &= \arg \min_\mathcal{E} |\mathcal{E}| \nonumber \\
    s.t. \qquad & C(\pi', \human + \mathcal{E}) \geq C^*_{\robot} \qquad\qquad \forall \pi' \in \hat{\Pi}^H \nonumber
\end{align}
\end{defn}

{In addition to minimal explanations, we will refer to a set of model updates to be a valid explanation if it renders the foil set to be suboptimal compared to the robot plan.
In this work, we will focus on case (1) above, a more constrained version of this explanation, namely one that establishes the invalidity of the foils. We will denote this explanation as $\mathcal{E}^{con}_{\text{VAL}}$. In this case we will require that after including the model updates none of the plans in $\hat{\Pi}^H$ is valid, i.e,
\[\qquad C(\pi', \human + \mathcal{E}^{con}_{\text{VAL}}) = \infty \qquad\qquad \forall \pi' \in \hat{\Pi}^H\]
Our focus on validity aligns with our basic setting, where human is the driver of the decision-making process and may have hidden preferences. This means, we don't want to disallow suboptimal plans the human may prefer, though our system could easily be extended to enforce optimality or consider cost thresholds that restricts how costly the human specified alternative can be.
}
{In our case, we are not directly given $\hat{\Pi}^H$, but rather the foil provided takes the form of a partial plan $\bar{\pi}^H$.
A partial plan can be defined as a tuple $\bar{\pi}^H = \langle \bar{A}, \prec \rangle$, where $\bar{A}$ is a multi-set consisting of actions and $\prec$ is a precedence relation between the actions in $\bar{A}$. A plan is said to be consistent with a partial plan (denoted as $\pi \models \bar{\pi}^H$), if the actions in $\bar{A}$ appears in $\pi$ and their precedence relations are satisfied by the order in which they appear. We expect each plan $\pi \in \hat{\Pi}^H$ to satisfy the provided $\bar{\pi}^H$. Our explanatory objective now becomes to find the minimal set of model updates $\bar{\mathcal{E}}^{con}_{\text{VAL}}$ such that there exists no plan in the updated model that satisfies the partial plan
\[\qquad \not \exists \pi', C(\pi', \human + \bar{\mathcal{E}}^{con}_{\text{VAL}}) \not= \infty \wedge \pi' \models \bar{\pi}^H\]}%
{Note that the minimal set $\bar{\mathcal{E}}^{con}_{\text{VAL}}$ can be larger than  $\mathcal{E}^{con}_{\text{VAL}}$ as the partial plan may have completions not part of $\hat{\Pi}^H$.
However, given that the system only knows $\bar{\pi}^H$, $\bar{\mathcal{E}}^{con}_{\text{VAL}}$ is the minimal explanation it can identify; we are guaranteed that $\bar{\mathcal{E}}^{con}_{\text{VAL}}$ will be valid as long as $\hat{\Pi}^H$ is a subset of the set of all plans that are consistent with $\bar{\pi}^H$.
One way to operationalize this would be to create a constrained planning model that only allows for solutions that satisfy $\bar{\pi}^H$. For partial plans, one could do this by directly using the compilation used in \cite{ramirez2009plan}. For a given model $\mathcal{M}$ and a partial foil $\bar{\pi}^H$, the compilation creates a new model $\mathcal{M}_{\bar{\pi}^H}$, such that the set of plans valid for $\mathcal{M}_{\bar{\pi}^H}$ is equal to the set of plans valid in $\mathcal{M}$ that satisfy the partial plan $\bar{\pi}^H$. This means we can now reframe our explanation objective as one of identifying a set of model updates that renders the compiled model unsolvable, i.e., we want to find $\bar{\mathcal{E}}^{con}_{\text{VAL}}$ such that
\[\qquad \not \exists \pi',~ C(\pi', \human_{\bar{\pi}^H} + \bar{\mathcal{E}}^{con}_{\text{VAL}}) \not= \infty \qquad\qquad \]%
where $\human_{\bar{\pi}^H}$ is the compiled human model.}

{To generate such explanations, \radarnew~adapts the MCE search in \cite{chakraborti2017plan} to satisfy this objective and come up with the required explanation. The goal test now checks for the unsolvability of the updated compiled model, by using a complete planner. Note that this can be an expensive algorithm given we call a planner at each step of the search. A way to speed up the process would be to use faster unsolvability tests that avoid the use of a planner. One possibility in particular is to use semi-relaxed reachability heuristics like $h^m$ \cite{geffner2000admissible} that are admissible and guaranteed to be equal to $h^*$ as $m$ tends to $|F|$. We can do this iteratively, where we start with $m=1$ and try to find a set of model updates $(\bar{\mathcal{E}}^{con}_{\text{VAL}})$ with
\[\qquad h^{m=i}(\hat{I}, \human_{\bar{\pi}^H} + \bar{\mathcal{E}}^{con}_{\text{VAL}})=\infty\]%
where $\hat{I}$ is the initial state in the updated model. So effectively for each value of $m$, we will try to search for a set of model updates for which the heuristic function says the goal is no longer reachable from the initial state. For lower values of $m$, this test can be performed relatively efficiently (with $m=1$ turning into testing for reachability in a delete-relaxed model). We will refer to this modified search as {\tt Approx-}MCrE Search. We can now easily establish the following properties of the search
\begin{prop}
{\tt Approx-}MCrE Search is \textbf{sound}, i.e., any explanation found by {\tt Approx-}MCrE is valid, and \textbf{complete}, i.e., it is guaranteed to find a valid explanation when one exists.
\end{prop}
\begin{prop}
Explanations generated through {\tt Approx-}MCrE Search need not be minimal.
\end{prop}
The first property follows from the fact that when $ h^{m}(\cdot)$ returns $\infty$ the problem must be unsolvable and for every unsolvable problem $ h^{m}(\cdot)$ will return $\infty$ for a high enough value of $m$. The second property follows from the fact that for some low value of $m$, a larger explanation may render the compiled problem unsolvable, while a smaller explanation may have been identified as valid for larger values of $m$. 
}

The explanation $\bar{\mathcal{E}}^{con}_{\text{VAL}}$ here can be viewed as the correction that needs to be made in the human model for refuting the suggested foil. In the other case where the specified foil is suboptimal or costlier than the suggested plan, we provide a message on the cost difference and allow the user to either enforce the suboptimal plan or ask for a model reconciliation based explanation. 
\begin{figure}[t]
    \centering
     \includegraphics[keepaspectratio=true, width=\columnwidth]{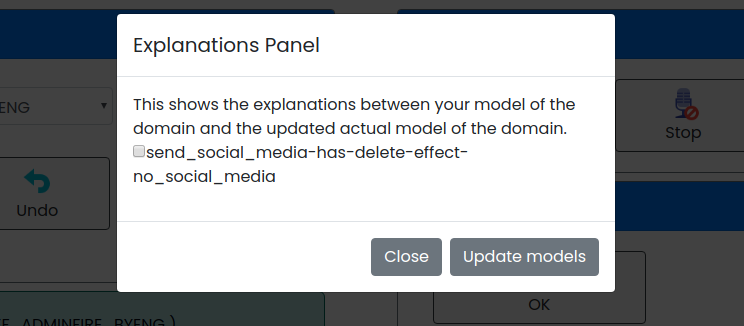}
    \caption{The minimally complete contrastive explanation generated.}
    \label{fig:exp}
\end{figure}

{For the use-case described above, \radarnew, given the foil, generates an  explanation where the action {\small SEND\_SOCIAL\_MEDIA\_BYENG\_BYENG} deletes an effect {\small NO\_SOCIAL\_MEDIA} which is a precondition of {\small ADDRESS\_MEDIA\_FIRECHIEF} (see Figure \ref{fig:exp}). Hence, the given foil is invalid in the planner's model. This explanation is the minimal explanation required to refute the foil. Once the explanation is presented, the human's model can then be corrected by adding the delete effect into the model.}


\subsection{Proactive Preference Elicitation-- Suggesting Plans}
Even though the human's model is updated and the human understands that the given foil is invalid, the foil serves a second purpose-- it is indicative of some latent preferences of the user. We hypothesize that asking for contrastive explanations exposes some of the preferences that the user does not specify explicitly. Thus, one can use foils to identify plans that are closer to the human's expectations. 
We use three different approaches in \radarnew~to identify such plans.

\subsubsection{The Closest Plan approach}
In this approach, we look at generating the {\em closest plan} to the specified foil which implies using the largest part of the foil in the revised plan. 
For this, we revisit the plan-recognition-as-planning methodology presented in \cite{ramirez2009plan} and construct a simple yet effective compilation that encodes the partial foils as soft constraints and imposes penalties if any of them are violated when coming up with a plan. This is similar in spirit to \cite{sohrabi2016plan}.
 \begin{figure}[t]
    \centering
     \includegraphics[keepaspectratio=true, width=\columnwidth]{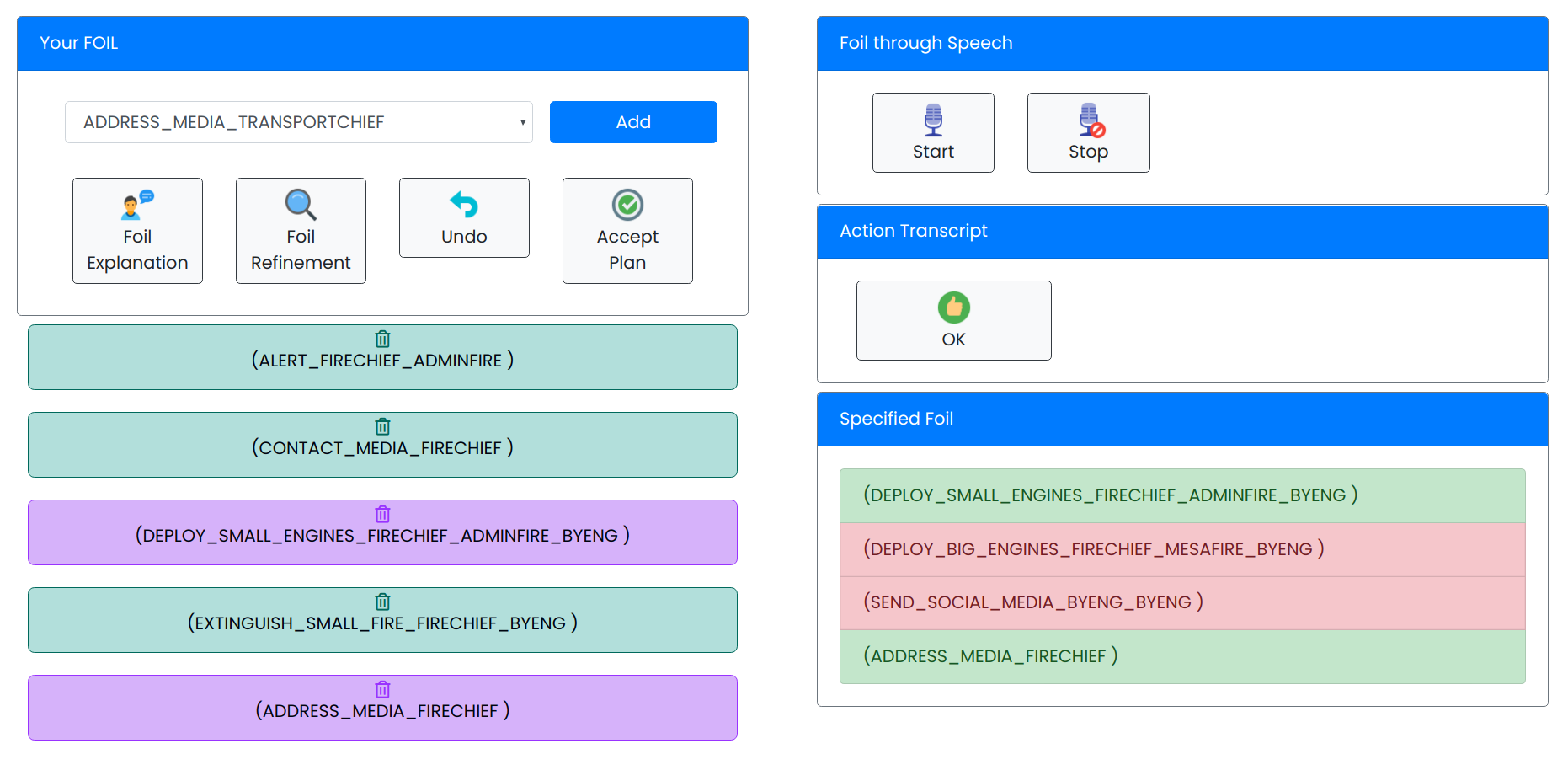}
    \caption{{Generating the closest plan that uses the largest part of the foil. }}
    \label{fig:nearestplan}
\end{figure}

A \textbf{Plan Recognition Problem} is represented by a tuple $R = \langle \mathcal{D},I,O,\mathcal{G}\rangle$, where $\mathcal{D}=\langle F,A \rangle$ is the planning domain, $I \subseteq F$ is the initial state, $\mathcal{G}$ is the set of possible goals $G$, $G \subseteq F$, and $O = \langle o_1, o_2,...,o_n \rangle$ is an observation sequence with each $o_i$ being an action in $A$ and $i \in [1,n]$.
We compile the foil to a new planning problem by augmenting the existing actions in the original problem with a set of `explain' and `discard' actions for each observation $o_i$ present in the observation sequence $O$. Here, the observations represent actions in the partial foil that the user specifies. An `explain' action for an observation $o_i$ is a replica of the observation with an additional effect ($\textit{met}_{o_i}$) that indicates the observation is met (i.e. action $o_i$ in the foil is used). On the other hand, a `discard' action for an observation $o_i$ is a dummy action which only has the effect ($\textit{met}_{o_i}$) but has a cost that is significantly higher than the corresponding `explain' action. This means discarding any observation should be costlier. For now, we look to preserve the ordering constraints of the observations by adding $\textit{met}_{o_{i-1}}$ as a precondition for the observation $o_i$ with $i \in [2,m]$.

\begin{defn}
\noindent A \textbf{Transformed Planning Problem} for a plan recognition problem $R = \langle \mathcal{D},I, O,\mathcal{G} \rangle$ is $P' = \langle \mathcal{D}',I',G'\rangle$  where $\mathcal{D}'=\langle F',A' \rangle $ and:
\begin{itemize}
    \item $F' = F \cup \{\textit{met}_{o_i}|o_i \in O\}$,
    \item $I' = I$
    \item $G' = \{g| \forall g \in \mathcal{G}\} \cup \{\textit{met}_{o_i}|o_i \in O\}$
    \item $A' = A \cup A_{explain} \cup A_{discard}$
    \begin{itemize}
        \item[$-$] $A_{explain} = \{e_{o_i}|o_{i}\in O,~c_{e_{o_i}} = c_{o_{i}},~\textit{pre}(e_{o_i}) = \{ \textit{pre}(o_{i}) \cup \textit{met}_{o_{i-1}}$ if $i>1;$ else, $\textit{pre}(o_{i})\},~\textit{eff}^{+}_{e_{o_i}} =\{ \textit{eff}^{+}_{o_{i}} \cup \textit{met}_{o_i}\},~ \textit{eff}^{-}_{e_{o_i}} = \{ \textit{eff}^{-}_{o_{i}} \}\}$
        \item[$-$] $A_{discard} = \{ d_{o_i}|o_{i}\in O,~c_{d_{o_i}}>>c_{e_{o_i}},~\textit{pre}(d_{o_i}) = \{\emptyset\},~\textit{eff}^{+}_{d_{o_i}}=\{\textit{met}_{o_i}\},~\textit{eff}^{-}_{e_{o_i}} = \{\emptyset\} \}$
    \end{itemize}
\end{itemize}
\end{defn}%
Using this transformed planning problem, the planner generates a plan that uses the largest possible part of the foil. 

{In the previously mentioned example, the generated plan contains of two actions present in the partial foil $\pi'$. The used actions {\small ADDRESS\_MEDIA\_FIRECHIEF} and {\small DEPLOY\_SMALL\_ENGINES\_FIRECHIEF\_ADMINFIRE\_BYENG} are encoded in a different color to help the user easily identify parts of the foil used (see Figure \ref{fig:nearestplan}). }
Additionally, the interface replaces \textit{`Current Plan'} panel with the \textit{`Specified Foil'} panel (at the bottom right of Figure \ref{fig:nearestplan}) where the actions used (observations met) in the generated plan are encoded in green color and actions discarded (unmet observations) are encoded in red. These actions can be added or deleted by clicking on the action in the panel. Further, we allow the users to directly add preferred actions that are not present in the generated plan, specify another foil, and engage in a longitudinal interaction. 

 \begin{figure}[t]
    \centering
     \includegraphics[keepaspectratio=true, width=\columnwidth]{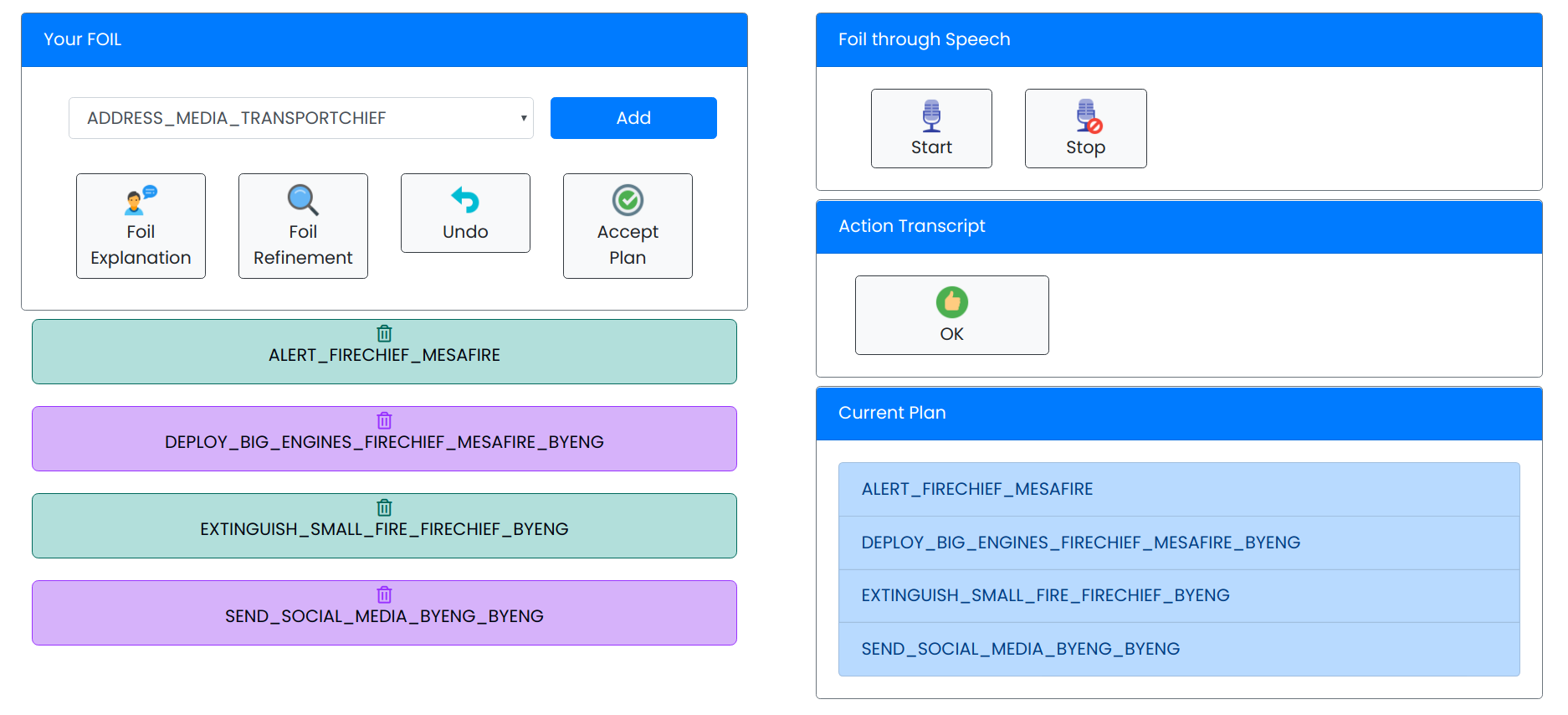}
    \caption{{Conflict-set resolution to generate preferred plans.}}
    \label{fig:conflict}
\end{figure}

\subsubsection{The Conflict Sets approach}
Even though the plan generated using the above compilation utilizes the largest part of the foil, the actions may have different importance to the user; hence, a planner may choose to use parts of the foil that are less important to the user. 
For example, the two actions from the specified foil ({\small ADDRESS\_MEDIA\_FIRECHIEF} and {\small DEPLOY\_SMALL\_ENGINES\_FIRECHIEF\_ADMINFIRE\_BYENG}) that were present in the generated plan may be less preferred by the user than the actions that were discarded ({\small SEND\_SOCIAL\_MEDIA\_BYENG\_BYENG} and {\small DEPLOY\_BIG\_ENGINES\_FIRECHIEF\_MESAFIRE\_BYENG}). 
Thus, to reach the final plan that the user prefers, they might have to engage in recurring interactions. This increases the amount of effort the user has to put in to make sure that the planner generates a plan of his/her liking. A simple attempt to reduce the cognitive load on the user would be to provide all possible sets of conflicting actions in the specified foil and ask the user to resolve them.
To find such conflict sets, we employ a systematic breadth-first search in the space of subsets of the foil. The idea here is similar to that of the Systematic Strengthening (SysS) approach in \cite{eifler2020new}. Starting from the empty set, for each subset of the specified foil, we use the compilation specified in \cite{ramirez2009plan} to compile it into a planning problem and check whether the compiled problem is unsolvable.\footnote{If we forget the need to get optimal conflict sets, we can rely on faster unsolvability tests  similar to the one mentioned previously.}
Subsets corresponding to unsolvable problems are presented to the user to resolve them by removing an action from the set. As mentioned earlier, we look to preserve the ordering constraints for now, hence, we do not consider the permutations of a subset; but this can be relaxed as well.

Once the conflict sets have been resolved by the user, the system can generate a plan that contains the preferred actions specified in the foil (which are non-conflicting) along with actions that are not part of a conflict set. While this method helps to elicit the user's preferences, the space of the action subsets derived from the foil, even barring permutations, explodes combinatorially. Thus, calculating the conflict sets can be expensive.

{In our use-case, \radarnew~searches all the subsets of the partial foil, starting from the empty set. {\small (DEPLOY\_SMALL\_ENGINES\_FIRECHIEF\_ADMINFIRE \_BYENG, DEPLOY\_BIG\_ENGINES\_FIRECHIEF\_MESAFIRE\_BYENG)} is a conflict set as the predicate {\small NO\_ENGINES\_DEPLOYED} gets deleted by the former action and is required as a precondition for the latter one. This is presented to the user to elicit the user's preferred action. Then, the next conflict set presented to the user is {\small(SEND\_SOCIAL\_MEDIA\_BYENG\_BYENG, ADDRESS\_MEDIA\_FIRECHIEF)}. Using both the preferences the final plan is generated and the preferred actions are encoded in a different color for the user to identify.\footnote{For our example, in the unsolvability tests with the $h^m$ pre-processor, m=1 found the solution.}
This is shown in Figure \ref{fig:conflict}.
Note that some of the conflict sets presented to the user may not be resolvable in the human's model without additional explanations. As a next step, we look to allow for providing such explanations.} 
\begin{figure}[t]
    \centering
     \includegraphics[keepaspectratio=true, width=\columnwidth]{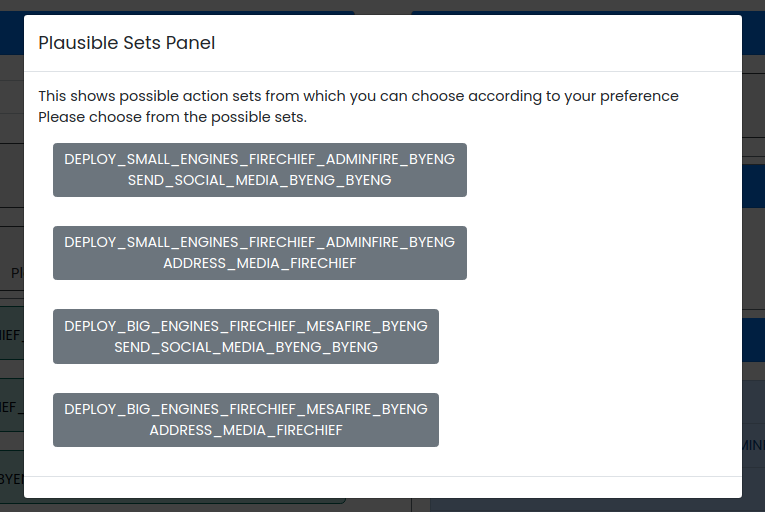}
    \caption{{Presenting maximal plausible sets to generate preferred plans.}}
    \label{fig:plausible}
\end{figure}
\subsubsection{The Plausible Sets approach} Instead of making the user resolve conflict sets, the system can also present the plausible subsets of the foil as options for the user to choose from. In this approach, we aim to present all the maximal valid subsets of the foil to the user. Maximal valid subsets can be considered as subsets which contain the maximum number of actions from the foil and a plan can be generated using \textit{all} of the actions present in the subset. To find such sets, we use an idea similar to that of Systematic Weakening (SysW) mentioned in \cite{eifler2020new}. Starting from the entire foil as a set, for each subset, we compile it into a planning problem (using the compilation specified in \cite{ramirez2009plan}) and try to generate a plan. An unsolvability test (similar to that of the previous approach) is done before generating the plan to discard subsets that are found to be unsolvable. In the case of successful plan generation, the corresponding subset is deemed to be valid. Note that subsets that are already part of a valid subset are not checked as we aim to present the maximal plausible sets. Once all the valid subsets are found, they are presented to elicit the preference of the user and then, based on the preference, a plan is suggested. Similar to the conflict sets approach, we preserve the ordering constraints but even then, this would be an expensive computation as the search is in the space of subsets.
{\radarnew~presents all the maximal plausible subsets for the previously illustrated example (as shown in Figure \ref{fig:plausible}) from which the user can choose the preferred one. For instance, if the user chooses the subset ({\small DEPLOY\_BIG\_ENGINES\_FIRECHIEF\_MESAFIRE\_BYENG} and {\small ADDRESS\_MEDIA\_FIRECHIEF}), a plan that contains both these actions is presented to the user.} 
\section{Evaluation}
\subsection{Demo Video} 
Using the fire-fighting scenario proposed in \cite{grover2020radar}, we illustrate the use cases and the functionalities supported by \radarnew~in a demo video. The demo video  of the working system can be found at \url{https://bit.ly/2Uzhciq}.
\subsection{User Study}
Here, we look at an evaluation for a major aspect of \radarnew~i.e., providing contrastive explanations. 
There have been existing works that provide algorithms for contrastive explanations in sequential decision-making settings, but these works take it as a given that users do ask for contrastive explanations with explicit foils. Further, in decision support scenarios, there hasn't been, to the best our knowledge, any studies that have looked at whether users would raise explicit foils when asking for an explanation. In this regard, we will now look at a user study we conducted that tries to answer the question of whether people ask for contrastive explanations with foils in decision support scenarios.
\subsubsection{Setup}
A simple logistics domain was used in this study and the task of a participant was (1) to understand the domain, (2) choose the right plan for a problem instance, and (3) look at the system's suggestion and evaluate it. To ensure users understood the domain, a questionnaire about the domain was given to the participants and they were only allowed to proceed after answering it correctly. Further, the participants get to chose what they think is the right plan before looking at the system's suggestion and for the purpose of this study, the system's suggested plan was made sure to be invalid in the model description that is exposed to the user.  The explanation for this invalidity constitutes of the system providing model information to the user which were previously withheld from them.
Once the plan was suggested, the participants had the option to either accept the suggested plan or ask for an explanation. If the participants asked for an explanation, they were presented with two explanatory questions as options-- (1) why did the system suggest the plan it suggested ($\pi^r$) as opposed to all other plans (in this case, Minimally Complete Explanations (MCE) were provided), or (2) why did the system not suggest the plan ($\pi^h$) that the participant chose (for which Minimally Contrastive Explanations (MCrE) were provided). Based on the explanatory question they choose, they were first provided with the respective explanation. Then, they were asked if they would accept the plan or if further explanations were required.
{Note that the paper aims to present the generalized system (which can handle general foils) to the end-users. But for the purpose of this user study, we had used full plans to compute MCrE. Since part of the motivation for providing contrastive explanations is the fact that the foil is something that the user expects or understands, we used the plan selected by the participants during the priming process as the foil to generate MCrE.}

In this study, there were two model differences that were required to explain the system's suggestion. MCE reconciled both the differences that were required to understand the plan. Therefore, when the participant who received MCE, asked for further explanations, no additional explanation was provided.
On the other hand, MCrE focused on one of the differences that made the participant's chosen plan invalid. In that case, when the participant asked for an additional explanation, the second difference that had not been reconciled was presented.
If the participant asked for further explanation after the additional explanation, no additional explanation was provided.

\subsubsection{Results and Analysis}
\begin{figure}[t]
    \centering
    \begin{tikzpicture}
        \pie[
        radius=1.5,
        explode=0.07,
        color={
            green!30!white,
            teal!30!white
        }]{
            38/Why not $\pi^h$?,
            62/Why $\pi^r$?
        }
    \end{tikzpicture}
    \caption{Preferences over explanatory questions}
    \label{fig:userstudy}
\end{figure}
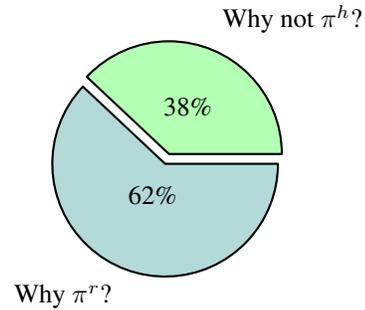
\begin{figure}[t]
    \centering
    \footnotesize
    \begin{tikzpicture}
        \begin{axis}[
            xbar stacked,
            height=4cm,
            width=7.5cm, 
            axis x line*=left,
            y axis line style = { opacity = 0 },
            x axis line style = { opacity = 0 , draw = none},
            enlarge y limits  = 0.7, 
            enlarge x limits  = 0.02,
            symbolic y coords = {MCE,MCrE,MCrE+Add~Exp},
            xlabel={\%},
            ytick=data,
             legend style={
                at={(rel axis cs:0.5,1)},
                anchor=south,
                legend columns=-1, 
                column sep=2mm, 
                draw=none 
            }
        ]
            \addplot[teal!30!black,fill=teal!95!black] coordinates { (42,MCE) (0,MCrE) (83,MCrE+Add~Exp)};
            \addplot[green!30!white,fill=green!30!white] coordinates {(58,MCE) (100,MCrE) (17,MCrE+Add~Exp)};
            \legend{Accepted, Not Accepted}
        \end{axis}
    \end{tikzpicture}
    \caption{Acceptance rate of the plan suggestion for each of the explanation type}
    \label{fig:userstudy2}
\end{figure}
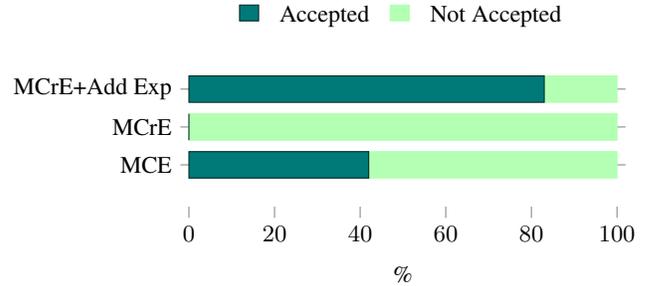
There were 35 students (undergraduate and graduate) who were the participants of this study. All the participants were able to pick the right plan for the problem instance. Out of these 35 participants, 32 participants asked for explanations. Of the 32, 12 (38\%) of them asked for MCrE and 20 (62\%) asked for MCE (as shown in Figure \ref{fig:userstudy}). This actually shows that a non-trivial number of participants had asked for a contrastive explanation. Note that the participants did not really have a stake in the domain and had to be primed in the study to get strong expectations. In real-life mission planning scenarios, where there are much more complex domains, the scientists, who are experts, have a stake in the domain and are highly likely to have strong expectations or preferences for one objective over another \cite{smith2012planning}. In such situations it could be even more likely that the users will ask for contrastive explanations. This emphasizes the need to provide contrastive explanations based on user specified foils.
\subsubsection{An interesting insight}
An interesting preliminary insight that was obtained in this study was that a higher percentage of the participants accepted the final plan when the explanations were broken down ($83\%$), i.e. they asked for MCrE and then an additional explanation compared to the participants who received both the model differences all at once in the case of MCE ($42\%$). This is shown in Figure \ref{fig:userstudy2}. This shows that providing information all at once (even if relevant) might not help the user comprehend the information properly. 
\subsection{Computational Evaluation}
The conflict sets approach and the plausible sets approach are searches in the space of subsets. This, as previously mentioned, can combinatorially explode as the number of actions in the foil increase. We have evaluated the time taken for both approaches on the firefighting scenario illustrated with two foils, one containing a single conflict set and the other containing 2 conflict sets (or $\pi'$). Table \ref{tab:my_label} shows the computation time taken by both the approaches on an Intel Xeon E3-1535 CPU equipped with 16GB RAM.  {Further, Table \ref{tab:my_label} also showcases the time taken for the system to generate MCrE for the given foil ($\pi'$) in the previously illustrated usecase. The computation time is feasible in the context of the application at hand. Even in the most computationally demanding setting, i.e., one with two foils the system only took around 30 seconds for generating revised suggestions. This is an acceptable computation time in mission critical decision-support scenarios, where the stakeholders usually deliberate over the final decision.}
\begin{table}[t]
    \centering
    \begin{tabular}{lcc}
    \toprule
 Approach & \multicolumn{2}{c}{Time (in secs)} \\ [0.1ex]
  & For 1 set & For $\pi'$\\
  \midrule
  Conflict Sets  &  8.34 & 30.34 \\ 
  Plausible Sets  & 9.12 & 29.16 \\ 
  MCrE & - & 51.03\\
  \bottomrule
    \end{tabular}
    \caption{Computation time for explanation generation and two of the preference elicitation approaches.}
    \label{tab:my_label}
\end{table}
\section{Related Work}
The spectrum of work in human-in-the-loop planning \cite{kambhampati2015human} ranges from the more traditional mixed-initiative settings \cite{ferguson1996trains,ai2004mapgen,10.5555/3298239.3298379} where the planners drove the interaction in these scenarios with the users `advising' them, to works in decision support systems \cite{grover2020radar,sengupta2018ma,mishra2019cap} where the user is responsible for the plan while the system provides support. In \cite{grover2020radar}, the authors propose a proactive aspect to decision support systems and design the system's capabilities based on principles in Human-Computer Interaction (HCI). In \cite{grover2019ipass}, the authors show that such systems can improve the efficiency of decision making, the quality of the decision made, and increase user-satisfaction. Our work builds upon existing work enabling the user to specify foils, refute them with explanation and engage in a discussion until consensus is reached. The explanations that our system provides take the human's model into account while other explainable tools focus on explaining the rationale behind the system's decision based on its own model \cite{agrawal2020using}.

\section{Conclusion}
In this article, we presented \radarnew, a decision-support system that looks to establish an interactive explanatory dialogue with the user. We looked at the two major technical aspects of this new interface. One enabled the user to ask for contrastive explanations by specifying a foil; this helps the user understand why the plan suggested by the system was chosen over the alternative. The other refined plan suggestions using the specified foil as a stand-in for the user's latent preferences. This was done by (1) providing a {\em closest plan} or (2) presenting {\em conflict sets} or (3) {\em maximal plausible sets} within the foil thereby eliciting the user's preferences. 

A crucial component in deploying decision support systems such as \radarnew~would be to have a specified model of the task given to the system. Once that is given, the system can offer all of its functionalities. Even explanations can be provided assuming an empty human model (as in \cite{grover2020radar}). We have developed this system with high-stakes mission-planning domains such as those in Navy or Space in mind, where there exists a specified model for the tasks that need to be carried out, with humans-in-the-loop having strong preferences over objectives. In such scenarios, the system deployed might have additional information as opposed to the human which causes a difference in the models. Also, the models could start off being the same but diverge during the course of operation. We believe \radarnew~would be of relevance and significant help in such domains.

\section{Acknowledgements}
{This research is supported in part by ONR grants N00014-16-1-2892, N00014-18-1-2442, N00014-18-1-2840, N00014-9-1-2119, AFOSR grant FA9550-18-1-0067, DARPA SAILON grant W911NF19-2-0006 and a JP Morgan AI Faculty Research grant. We thank Erin Chiou, the reviewers and the members of the Yochan research group for helpful discussions and feedback.}

\bibliography{aaai22.bib}

\begin{thebibliography}{22}
\providecommand{\natexlab}[1]{#1}

\bibitem[{Agrawal, Yelamanchili, and Chien(2020)}]{agrawal2020using}
Agrawal, J.; Yelamanchili, A.; and Chien, S. 2020.
\newblock Using explainable scheduling for the mars 2020 rover mission.
\newblock \emph{arXiv preprint arXiv:2011.08733}.

\bibitem[{Ai-Chang et~al.(2004)Ai-Chang, Bresina, Charest, Chase, Hsu, Jonsson,
  Kanefsky, Morris, Rajan, Yglesias et~al.}]{ai2004mapgen}
Ai-Chang, M.; Bresina, J.; Charest, L.; Chase, A.; Hsu, J.-J.; Jonsson, A.;
  Kanefsky, B.; Morris, P.; Rajan, K.; Yglesias, J.; et~al. 2004.
\newblock Mapgen: mixed-initiative planning and scheduling for the mars
  exploration rover mission.
\newblock \emph{IEEE Intelligent Systems}, 19(1): 8--12.

\bibitem[{Chakraborti et~al.(2017)Chakraborti, Sreedharan, Zhang, and
  Kambhampati}]{chakraborti2017plan}
Chakraborti, T.; Sreedharan, S.; Zhang, Y.; and Kambhampati, S. 2017.
\newblock Plan explanations as model reconciliation: Moving beyond explanation
  as soliloquy.
\newblock In \emph{Proc. IJCAI}.

\bibitem[{Eifler et~al.(2020)Eifler, Cashmore, Hoffmann, Magazzeni, and
  Steinmetz}]{eifler2020new}
Eifler, R.; Cashmore, M.; Hoffmann, J.; Magazzeni, D.; and Steinmetz, M. 2020.
\newblock A New Approach to Plan-Space Explanation: Analyzing Plan-Property
  Dependencies in Oversubscription Planning.
\newblock In \emph{AAAI}, 9818--9826.

\bibitem[{Ferguson et~al.(1996)Ferguson, Allen, Miller
  et~al.}]{ferguson1996trains}
Ferguson, G.; Allen, J.~F.; Miller, B.~W.; et~al. 1996.
\newblock TRAINS-95: Towards a Mixed-Initiative Planning Assistant.
\newblock In \emph{AIPS}, 70--77.

\bibitem[{Geffner and Haslum(2000)}]{geffner2000admissible}
Geffner, P. H.~H.; and Haslum, P. 2000.
\newblock Admissible heuristics for optimal planning.
\newblock In \emph{Proceedings of the 5th Internat. Conf. of AI Planning
  Systems (AIPS 2000)}, 140--149.

\bibitem[{Google(2017)}]{GoogleSTT}
Google. 2017.
\newblock Speech to Text API.
\newblock \url{https://cloud.google.com/speech-to-text}.
\newblock Accessed: 2021-10-30.

\bibitem[{Grover et~al.(2019)Grover, Sengupta, Chakraborti, Mishra, and
  Kambhampati}]{grover2019ipass}
Grover, S.; Sengupta, S.; Chakraborti, T.; Mishra, A.~P.; and Kambhampati, S.
  2019.
\newblock iPass: A Case Study of the Effectiveness of Automated Planning for
  Decision Support.
\newblock \emph{HCI Journal}.

\bibitem[{Grover et~al.(2020)Grover, Sengupta, Chakraborti, Mishra, and
  Kambhampati}]{grover2020radar}
Grover, S.; Sengupta, S.; Chakraborti, T.; Mishra, A.~P.; and Kambhampati, S.
  2020.
\newblock RADAR: automated task planning for proactive decision support.
\newblock \emph{Human--Computer Interaction}, 1--26.

\bibitem[{Kambhampati, Knoblock, and Yang(1995)}]{kambhampati1995planning}
Kambhampati, S.; Knoblock, C.~A.; and Yang, Q. 1995.
\newblock Planning as refinement search: A unified framework for evaluating
  design tradeoffs in partial-order planning.
\newblock \emph{Artificial Intelligence}.

\bibitem[{Kambhampati and Talamadupula(2015)}]{kambhampati2015human}
Kambhampati, S.; and Talamadupula, K. 2015.
\newblock Human-in-the-loop planning and decision support.
\newblock \emph{AAAI Tutorial}.

\bibitem[{Kim, Banks, and Shah(2017)}]{10.5555/3298239.3298379}
Kim, J.; Banks, C.~J.; and Shah, J.~A. 2017.
\newblock Collaborative Planning with Encoding of Users’ High-Level
  Strategies.
\newblock In \emph{Proceedings of the Thirty-First AAAI Conference on
  Artificial Intelligence}, AAAI’17, 955–961. AAAI Press.

\bibitem[{McDermott et~al.(1998)McDermott, Ghallab, Howe, Knoblock, Ram,
  Veloso, Weld, and Wilkins}]{mcdermott1998pddl}
McDermott, D.; Ghallab, M.; Howe, A.; Knoblock, C.; Ram, A.; Veloso, M.; Weld,
  D.; and Wilkins, D. 1998.
\newblock PDDL--the planning domain definition language--version 1.2.
\newblock \emph{Yale Center for Computational Vision and Control, Tech. Rep.
  CVC TR-98-003/DCS TR-1165}.

\bibitem[{Miller(2018)}]{miller2018contrastive}
Miller, T. 2018.
\newblock Contrastive explanation: A structural-model approach.
\newblock \emph{arXiv preprint arXiv:1811.03163}.

\bibitem[{Miller(2019)}]{miller2019explanation}
Miller, T. 2019.
\newblock Explanation in artificial intelligence: Insights from the social
  sciences.
\newblock \emph{Artificial Intelligence}.

\bibitem[{Mishra et~al.(2019)Mishra, Sengupta, Sreedharan, Chakraborti, and
  Kambhampati}]{mishra2019cap}
Mishra, A.~P.; Sengupta, S.; Sreedharan, S.; Chakraborti, T.; and Kambhampati,
  S. 2019.
\newblock Cap: A decision support system for crew scheduling using automated
  planning.
\newblock \emph{Naturalistic Decision Making}.

\bibitem[{Ram{\'\i}rez and Geffner(2009)}]{ramirez2009plan}
Ram{\'\i}rez, M.; and Geffner, H. 2009.
\newblock Plan recognition as planning.
\newblock In \emph{Twenty-First International Joint Conference on Artificial
  Intelligence}.

\bibitem[{Sengupta, Chakraborti, and Kambhampati(2018)}]{sengupta2018ma}
Sengupta, S.; Chakraborti, T.; and Kambhampati, S. 2018.
\newblock MA-RADAR--A mixed-reality interface for collaborative decision
  making.
\newblock \emph{ICAPS UISP}.

\bibitem[{Smith(2012)}]{smith2012planning}
Smith, D.~E. 2012.
\newblock Planning as an iterative process.
\newblock In \emph{Twenty-Sixth AAAI Conference on Artificial Intelligence}.

\bibitem[{Sohrabi, Riabov, and Udrea(2016)}]{sohrabi2016plan}
Sohrabi, S.; Riabov, A.~V.; and Udrea, O. 2016.
\newblock Plan Recognition as Planning Revisited.
\newblock In \emph{IJCAI}, 3258--3264.

\bibitem[{Sreedharan, Kambhampati et~al.(2018)}]{sreedharan2018handling}
Sreedharan, S.; Kambhampati, S.; et~al. 2018.
\newblock Handling model uncertainty and multiplicity in explanations via model
  reconciliation.
\newblock In \emph{ICAPS}.

\bibitem[{Sreedharan et~al.(2019)Sreedharan, Srivastava, Smith, and
  Kambhampati}]{sreedharan2019can}
Sreedharan, S.; Srivastava, S.; Smith, D.; and Kambhampati, S. 2019.
\newblock Why Can't You Do That HAL? Explaining Unsolvability of Planning
  Tasks.
\newblock In \emph{International Joint Conference on Artificial Intelligence}.

\end{thebibliography}
\end{document}